\documentclass{article}

\PassOptionsToPackage{numbers}{natbib}

\usepackage[final]{neurips_2022}

\usepackage[utf8]{inputenc} 
\usepackage[T1]{fontenc}    
\usepackage{xcolor}         
\usepackage{url}            
\usepackage{hyperref}

\usepackage{booktabs}       
\usepackage{amsfonts}       
\usepackage{amsmath,amssymb,amsthm}

\usepackage{nicefrac}       
\usepackage{microtype}      
\usepackage{multirow}

\usepackage{amsmath,amsfonts,bm}

{}
{}
{}
{}

\def\eqref#1{equation~\ref{#1}}

\def\1{\bm{1}}

\def\vg{{\bm{g}}}

\def\vk{{\bm{k}}}

\def\vv{{\bm{v}}}

\def\vx{{\bm{x}}}
\def\vy{{\bm{y}}}

\def\mA{{\bm{A}}}

\def\mG{{\bm{G}}}

\def\mR{{\bm{R}}}

\def\mW{{\bm{W}}}
\def\mX{{\bm{X}}}

\DeclareMathAlphabet{\mathsfit}{\encodingdefault}{\sfdefault}{m}{sl}
\SetMathAlphabet{\mathsfit}{bold}{\encodingdefault}{\sfdefault}{bx}{n}
\newcommand{\tens}[1]{\bm{\mathsfit{#1}}}

\def\tW{{\tens{W}}}

\usepackage{graphicx}
\usepackage{subfig}

\title{Learning to  Control Rapidly Changing Synaptic Connections: An Alternative Type of Memory in Sequence Processing Artificial Neural Networks}

\author{Kazuki Irie$^{1}$ ~ J\"urgen Schmidhuber$^{1,2}$\\
  $^1$The Swiss AI Lab, IDSIA, USI \& SUPSI, Lugano, Switzerland \\
  $^2$AI Initiative, KAUST, Thuwal, Saudi Arabia \\
  \texttt{\{kazuki, juergen\}@idsia.ch}
}

\begin{document}

\maketitle

\begin{abstract}
Short-term memory in standard, general-purpose, sequence-processing recurrent neural networks (RNNs) is stored as activations of nodes or  ``neurons.''
Generalising feedforward NNs to such RNNs is mathematically straightforward and natural, and even historical: already in 1943,
McCulloch and Pitts proposed this as a surrogate to ``synaptic modifications''
 (in effect, generalising the Lenz-Ising model, the first non-sequence processing RNN architecture of the 1920s).
A lesser known alternative approach to storing short-term memory in  ``synaptic connections''---by parameterising and controlling the dynamics of a context-sensitive time-varying weight matrix through another NN---yields another ``natural'' type of short-term memory in sequence processing NNs: the Fast Weight Programmers (FWPs) of the early 1990s. FWPs have seen a recent revival
as generic sequence processors, achieving competitive performance across various tasks.
They are formally closely related to the now popular Transformers. 
Here we present them in the context of artificial NNs as an abstraction of biological NNs---a perspective that has not been stressed enough in previous FWP work.
We first review aspects of FWPs for pedagogical purposes, then discuss connections to related works motivated by insights from neuroscience.
\end{abstract}

\section{Introduction}
\label{sec:intro}
Memory is essential for problem solving in the natural world in which presently available information may become important later.
In the context of artificial neural networks (NNs), we may distinguish two major categories of memory: long-term (LTM) and short-term memory (STM).
For example, a supervised feedforward NN (FNN) learning to solve some prediction task first sees a sequence of (batches of) training examples, then test examples.
During the training phase, its weight matrix (WM) is iteratively modified by some learning algorithm to reduce the NN's errors.
In the conventional setting,
the WM is frozen once training ends, and becomes a permanent form of memory (LTM) reflecting its training experience, repeatedly used to make predictions about unseen test examples. 
This LTM stored in the WM is the NN's program, where
the NN architecture is considered  a computer \citep{schmidhuber1990making}.\looseness=-1

When the world is only partially observable,
problem-solving NNs generally  need additional \textit{short-term memory} (STM), to temporally store information about previously observed inputs.
STM is typically specific to the current task, to be erased when the next task starts.
A general and standard sequence processing NN is the recurrent NN (RNN; \citep{jordan1986, elman1990finding}, see also older works reviewed in Sec.~\ref{sec:rel}).
While the transition from an FNN to an RNN is straightforward (reviewed in Sec.~\ref{sec:rnn}),
it is worth noting the underlying decision of storing STM as activations of ``neurons'' while keeping the WM, i.e., the ``synaptic connection weight strengths'' fixed.
Here our main goal is to shed light on an alternative way of introducing STM into the FNN: the Fast Weight Programmers (FWPs; \citep{Schmidhuber:91fastweights, schmidhuber1992learning}) which learn to store STM in synaptic connection weights.

\section{Short-Term Memory in Sequence Processing Nets}
\subsection{Conventional Approach: Storing Short-Term Memory in Neurons}
\label{sec:rnn}
Before presenting the alternative memory type in Sec.~\ref{sec:synap},
we first review ``conventional'' STM in sequence processing NNs.
In what follows, let $d_\text{in}$, $d_\text{out}$, and $t$ denote positive integers.
We define a layer of an NN as a function with a weight matrix $\mW \in \mathbb{R}^{d_\text{out} \times d_\text{in}}$ which transforms an input vector $\vx_t \in \mathbb{R}^{d_\text{in}}$ to an output vector $\vy_t \in \mathbb{R}^{d_\text{out}}$ as
\begin{eqnarray} \label{eq:ann}
\vy_t = \sigma(\mW \vx_t)
\end{eqnarray}
where $\sigma$ is an element-wise activation function.
We omit the additive bias term without loss of generality.
This is indeed an abstraction for an NN where $\vx$ and $\vy$ represent the activities of input and output ``neurons'' respectively, and the weight matrix $\mW$ stores strengths of ``synaptic connections'' between them.\footnote{
Generally speaking, the analogy never goes beyond this, and has limited (if any) practical implications or benefits.
}
This system has no STM: when we feed another input 
$\vx_{t+1}$ at the next time step $t+1$, the previous input $\vx_t$ has no influence on the computation of  output $\vy_{t+1}$. 
This \textit{feedforward} layer can be straightforwardly extended through an STM by making the output $\vy_t$ at step $t$ dependent on its own output $\vy_{t-1}$ from the previous time step $t-1$ by introducing another weight matrix $\mR \in \mathbb{R}^{d_\text{out} \times d_\text{out}}$
\begin{eqnarray}
\label{eq:rnn}
\vy_t = \sigma(\mW \vx_t + \mR \vy_{t-1})
\end{eqnarray}
This is a trivial and natural way of introducing STM to ANNs for sequence processing\footnote{See, e.g., \citet{elman1990finding}'s simplification of \citet{jordan1986}'s version featuring one more layer before the recurrence.}.
Essentially we model the dynamics of the output neurons $\vy_t$ over time by another fictive NN with 
synaptic connections $\mR$ between ``neurons at time $t$'' and ``neurons at time $t-1$''.
Since no such synaptic connections over time exist physically, we may be implicitly giving up the analogy to biological NNs here.
It is also worth noting that, while \citet{jordan1986} and \citet{elman1990finding} propose this kind of recurrence purely for the purpose of introducing temporal dependency, work by \citet{mcculloch1943logical} in 1943 already proposed it as a replacement of modifiable synaptic connections---see \textit{``Theorem 7. Alterable synapses can be replaced by circles.''}
Now what if we modelled the ``\textit{alterable synapses}'' directly? The alternative memory type presented in the next section represents an answer to this question.

\subsection{Alternative Approach: Storing Short-Term Memory in Synaptic Connections}
\label{sec:synap}
Restarting from Eq.~\ref{eq:ann},
we now show how to introduce STM to FNNs by explicitly and rapidly modifying the synaptic weights as a function of the inputs.
The core idea is to control the mechanism of synaptic weight changes by another learning NN.
Similarly to the recurrent connections $\mR$ introduced in the standard RNN (Sec.~\ref{sec:rnn}), this NN may be ``fictive'' from the perspective of the biological NN analogy.
We present a bottom-up step-by-step construction as follows.
Starting from the goal, we want $\mW$ to be parameterised as a function of the inputs,
i.e., we want $\mW_{t-1}$ to become $\mW_t$ at each step $t$.
This requires an update rule.
For now, let us take a very simple update equation: 
\begin{eqnarray}
\label{eq:up}
\mW_t = \mW_{t-1} + \vv_t \otimes \vk_t
\end{eqnarray}
where $\otimes$ denotes outer product $\vv_t \otimes \vk_t \in \mathbb{R}^{d_\text{out} \times d_\text{in}}$ between two vectors $\vv_t \in \mathbb{R}^{d_\text{out}}$ and $\vk_t \in \mathbb{R}^{d_\text{in}}$.
This is a generic equation akin to Hebb's informal learning rule \citep{hebb1949organization} ($\vk_t$ is the input and $\vv_t$ the output of this layer) as well as to the gradient descent update rule of the WM of a linear layer ($\vk_t$ as the input and $\vv_t$ the gradient of the loss w.r.t. the linear layer's output scaled by a negative learning rate).
Now, these newly introduced variables $\vk_t$ and $\vv_t$ have to be
generated somehow.
We can simply generate them from the actual external input $\vx_t$, i.e.,
\begin{eqnarray}
[\vk_t, \vv_t] = \mA \vx_t
\end{eqnarray}
where $\mA \in \mathbb{R}^{(d_\text{in} + d_\text{out}) \times d_\text{in}}$ is a weight matrix, and the square brackets denote vector concatenation.
Putting these operations in the right order yields a sequence processor that, just like the standard RNN, transforms an input $\vx_t \in  \mathbb{R}^{d_\text{in}}$ into an output $\vy_t \in \mathbb{R}^{d_\text{out}}$ at each time step $t$ as follows
\begin{eqnarray}
\label{eq:sl}
[\vk_t, \vv_t] &=& \mA \vx_t\\
\mW_t &=& \mW_{t-1} + \vv_t \otimes \vk_t \label{eq:weight_sum}\\
\vy_t &=& \sigma(\mW_t \vx_t) \label{eq:synap}
\end{eqnarray}
Essentially, an NN with parameters $\mA \in \mathbb{R}^{(d_\text{in} + d_\text{out}) \times d_\text{in}}$ translates the input observations into weight changes through Eqs.~\ref{eq:sl}-\ref{eq:weight_sum}.
This effectively augments Eq.~\ref{eq:ann} of an FNN with STM that is stored in the context-sensitive weight matrix $\mW_t$ representing the dynamic synaptic weights.
This idea is what was introduced by \citet{Schmidhuber:91fastweights, schmidhuber1992learning} as \textit{``an alternative to recurrent nets''}\footnote{Up to slight changes introduced for pedagogical purposes: we placed an activation function in Eq.~\ref{eq:synap} instead of Eq.~\ref{eq:weight_sum} to facilitate direct comparison to the original FNN of Eq.~\ref{eq:ann}.}.
The original work presents this model as a system of two networks:
the network in Eq.~\ref{eq:synap} is the \textit{fast net}, whose \textit{fast weights} $\mW_t$ are modified at every time step by the \textit{slow net} of Eq.~\ref{eq:sl} whose \textit{slow weights} $\mA$ are trained e.g, by backpropagation through time.
In the original paper \citep{Schmidhuber:91fastweights, schmidhuber1992learning}, the slow net is said to ``control'' the fast weights (and thus, it is a ``fast weight controller''), while recent works refer to it as a ``Fast Weight Programmer" (FWP) \citep{schlag2021linear};
considering the WM as the program of an NN \citep{schmidhuber1990making}. 
Modifying the WM in a goal-oriented manner means programming the NN via fast weight changes.

\paragraph{Practical Enhancements.}
Two technical enhancements have improved FWPs in practice.
Firstly, Transformers with linearised self-attention \citep{katharopoulos2020transformers, choromanski2020rethinking} have a ``dual form'' that is an outer product-based FWP:
exactly the same mathematical relation \citep{irie2022dual} maps the perceptron to its dual form, the kernel machine \citep{aizerman1964theoretical}.
A practical implication is that
FWPs can directly adopt advancements of architectural designs originally developed for Transformers \citep{trafo}, as FWPs implement their (linearised) self-attention layers.
These designs include an additional \textit{query} projection that generates the input to the fast net, as well as the now standard two-layer feedforward blocks and layer normalisation.
In fact, it is now also common to use the associative memory terminology  ``key/value''  to describe FWPs (as reflected in the notations $\vk_t$ and $\vv_t$ above), rather than the original ``FROM/TO'' terminology of the 1990s.\looseness=-1

The second kind of enhancement is an improved update rule.
While the ``purely additive'' update rule of Eq.~\ref{eq:up} is the one used in the standard Linear Transformer \citep{katharopoulos2020transformers},
\citet{schlag2021linear} identify its memory capacity problem and proposes to replace it by the delta learning rule \citep{widrow1960adaptive, schlag2020fastweightmemory}, which has been shown to consistently outperform the purely additive one on various tasks including language modelling \citep{schlag2021linear}, algorithmic tasks \citep{irie2021going}, time series prediction \citep{irie2022neural}, image generation \citep{irie2022image}, and video game playing in reinforcement learning \citep{irie2021going}.

\paragraph{Properties.}
The incremental/additive nature of STM updates in FWPs (Eq.~\ref{eq:weight_sum}) yields interesting properties that the standard RNN does not have.
Firstly, it provides good gradient flow \citep{schmidhuber2021fwp} and alleviates the fundamental vanishing gradient problem \citep{hochreiter1991untersuchungen} of sequence learning through gradient descent.
Second, unlike in standard RNNs, the weight update equation (e.g., Eq.~\ref{eq:weight_sum})
can be directly seen as an Euler discretisation of its continuous-time counterpart.
This can be exploited to derive a powerful Neural-ODE/CDE based sequence processors for the continuous-time domain \citep{irie2022neural}.

Another important property not shared by standard RNNs is the number of temporal variables, one of the original motivations of FWPs \citep{schmidhuber1993reducing}.
Assuming $d=d_{\text{in}}=d_{\text{out}}$, FWPs store $O(d^2)$ memory
instead of $O(d)$ in the case of RNNs.
While this may increase the memory storage, it also  increases the space complexity of FWPs trained by backpropagation through time.
A naive implementation has a space complexity of $O(T d^2)$  where $T$ is the number of backpropagation steps.
Fortunately, the incremental/additive nature of memory allows for deriving memory-efficient backpropagation that reduces this complexity to $O(d^2)$ (see e.g., \citep{katharopoulos2020transformers, schlag2021linear, IrieSCS22}), allowing for scaling up practical applications.

\paragraph{Other Extensions.}
There is no reason to restrict the STM storage to either ``neurons'' or ``synaptic weights.''
We can use both:
one can parameterise weight matrices in the standard RNN (Eq.~\ref{eq:rnn}) as a function of inputs using the FWP principle of Sec.~\ref{sec:synap}. Alternatively, one can also introduce recurrent connections to the FWP that connect activations of the fast net from the previous time step to the input of the slow net.
These hybrid approaches \citep{irie2021going} have been shown to improve over other FWPs without recurrence on various tasks.

Another natural extension is to make all weights context-dependent, including the slow weights in the FWP themselves.
Such a model can be obtained with minor modifications to Eqs.~\ref{eq:sl}-\ref{eq:synap} \citep{IrieSCS22}.
These self-referential variants \citep{Schmidhuber:92selfref, Schmidhuber:93selfreficann, schmidhuber1987evolutionary} allow an explicit formulation of recursive self-modifications in artificial NNs, another interesting perspective directly derived from the FWP principle.

\section{Relating to Works Motivated by Neuroscience}
\label{sec:rel}
The importance of rapidly changing NN weights was first explicitly emphasised by \citet{von1981correlation}, then also by other authors \citet{feldman1982dynamic, hinton1987using}.
The parameterisation and control of synapses through another NN, however,
was introduced only  in the early 1990s by \citet{Schmidhuber:91fastweights, schmidhuber1992learning}.
There are also several rather recent works motivated by insights from neuroscience to improve ``synapses'' in artificial NNs.
For example, \citet{LahiriG13} and \citet{ZenkePG17} question the
single-scalar parameterisation of synapses in artificial NNs in light of the complexity of biological synapses.
The core idea of FWPs directly relates to this spirit, as FWPs parameterise synapses by a general purpose NN that can potentially  control any complex dynamics.
The models of \citet{miconi18a, miconi2018backpropamine} motivated by synaptic plasticity in the brain (see also \cite{RodriguezGM22}) are FWP variants where plasticity is introduced as augmentation to some base RNN (in the spirit of the ``hybrid approaches'' mentioned above in Sec.~\ref{sec:synap} ``Other Extensions'').
Here we stress that such synaptic weight modifications can be a stand-alone mechanism for STM.
In machine learning, dynamic synapses are also motivated by applications in meta-learning (e.g., \citep{munkhdalai2017meta, munkhdalai2018metalearning, MunkhdalaiSWT19, NajarroR20, kirsch2021}) and continual learning (e.g., \citep{ZenkePG17, munkhdalai2020sparse, ThangarasaMT20}).

The work of \citet{WhittingtonWB22}
implicitly relates a model of hippocampal formation (Tolman-Eichenbaum Machine; TEM \citep{whittington2020tolman}) to Transformers without softmax in self-attention, which
 relate to the so-called Linear Transformers \citep{katharopoulos2020transformers}, which are FWPs \citep{schlag2021linear}.
Specifically, TEM (simplified with some reasonable assumptions \citep{WhittingtonWB22}) is an auto-regressive sequence processor that, at each time step $t > 0$, estimates the next sensory input $\vy_t \in \mathbb{R}^{d_\text{in}}$ as a function of the current sensory input $\vx_t \in \mathbb{R}^{d_\text{in}}$, the action represented by an integer $a_t$, and the positional representation $\vg_t \in \mathbb{R}^{d_\text{pos}}$ (where $d_\text{pos}$ is a positive integer) corresponding to activities of ``grid cells'' (or ``medial entorhical cells'').
$\vg_t$ is parameterised by an RNN without external inputs, with an action-dependent weight matrix $\tW_{a_t} = f(a_t) \in \mathbb{R}^{d_\text{pos} \times d_\text{pos}}$ where $f$ is a parameterised feedforward NN as follows:\footnote{We introduce two notational changes compared to \citet{WhittingtonWB22}. First, for consistency with the rest of this paper, we use column-major vectors.
Second, for clarity, we introduce time indices to all variables.
We also omit many projection layers including extra layer after $\vy_t$ as well as some additional residual connections that are in the official implementation.}
\begin{eqnarray}
\vg_t &=& \sigma(\tW_{a_t} \vg_{t-1}) \\
\vy_t &=& \alpha \mX_t \mG_t^\intercal \vg_t
\end{eqnarray}
where $\mX_t \in \mathbb{R}^{d_\text{in} \times t}$ and $\mG_t \in \mathbb{R}^{d_\text{pos} \times t}$ are matrices constructed by concatenating the corresponding vectors from the previous steps, i.e., $\mX_t = [\vx_1, ..., \vx_t]$ and $\mG_t = [\vg_1, ..., \vg_t]$, and $\alpha \in \mathbb{R}$.
Clearly, this computation can be expressed as an NN (linear layer) with context-dependent synaptic connections $\mW_t \in \mathbb{R}^{d_\text{in} \times d_\text{pos}}$ (with $\mW_0 =0$) that evolve over time in FWP fashion as follows:
\begin{eqnarray}
\mW_t &=&  \alpha \mX_t \mG_t^\intercal = \mW_{t-1} + \alpha \vx_t \otimes \vg_t \\
\vy_t &=& \mW_{t} \vg_t
\end{eqnarray}
Unlike in the original formulation of TEM, the activities of ``place cells'' do not directly appear in this FWP formulation (they are hidden in the computation of dynamic synaptic connection weights).
We also note that this connection is not surprising, since the motivation of TEM \cite{whittington2020tolman} (``structural generalisation'') is a form of systematic generalisation \citep{fodor1988connectionism}, and FWPs relate to tensor product representations that are popular in the context of systematic generalisation \citep{smolensky1990tensor, schlag2018learning}.

The general principle of storing patterns in LTM based on changing synaptic connections is a much older concept already found in non-learning (and non-sequence processing) RNNs.
In the 1970s \citet{amari1972learning} extended the Ising or Lenz-Ising model introduced in the 1920s \citep{lenz1920beitrvsge, ising1924thesis, ising1925beitrag, kramers1941statistics, wannier1945statistical} (where a binary neuron is analogous to a spin), by making it adaptive and thus capable of learning associations of input/output patterns by changing the connection weights. This precedes the work of \citet{little1974existence} and \citet{hopfield1982neural} (see also \citep{amit1985spin}) on what has been called the "Hopfield Network" or Amari-Hopfield Network \citep{millan2019memory} that has also been recently revisited \citep{ramsauer2021hopfield, KrotovH21, MillidgeS0LB22}.
\citet{amari1972learning} even discusses the sequence processing scenario.

As mentioned above, the principles of standard RNNs were also discussed by neuroscientists \citet{mcculloch1943logical} in 1943,
and analysed by \citet{kleene1956representation} in the 1950s.
\citet{turing1948}'s unpublished ``unorganized machines'' of 1948 also relate to RNNs.

\section{Conclusion}
We discussed Fast Weight Programmers (FWPs) as an alternative to standard RNNs or to McCulloch and Pitts' model with ``cycles'' that replace ``alterable synapses'' (see their informal Theorem 7).
This interesting perspective has not been stressed much in previous work on FWPs.
While the exact role of synaptic plasticity on memory in the brain remains unclear (see ongoing discussions of this topic in neuroscience, e.g., \citep{trettenbrein2016demise, gershman2021molecular}),
we saw how compactly and elegantly FWPs can implement the learnable dynamics of short-term memory in time-varying synaptic connections of artificial NNs.

\section*{Acknowledgements}
This research was partially funded by ERC Advanced grant no: 742870, project AlgoRNN,
and by Swiss National Science Foundation grant no: 200021\_192356, project NEUSYM.

\bibliography{references}
\bibliographystyle{unsrtnat}

\end{document}